\documentclass[journal]{IEEEtran}
\ifCLASSINFOpdf
\else
\fi
\usepackage{graphicx,tabularx,booktabs,makecell,xcolor}
\usepackage{subcaption}
\usepackage{hyperref}

\usepackage[english]{babel}
\usepackage[autostyle, english = american]{csquotes}
\MakeOuterQuote{"}

\usepackage{amsmath,amssymb,amsfonts,float}
\begin{document}
%
\title{Reinforcement Learning Enhancement Using Vector Semantic Representation and Symbolic Reasoning for Human-Centered Autonomous Emergency Braking}
%
%
%

\author{Vinal~Asodia,~Iman~Sharifi,~and~Saber~Fallah
\thanks{$^{1}$V. Asodia and S. Fallah are with CAV-Lab at the School of Engineering, University of Surrey, Guildford, GU2 7XH UK.
        {\tt\small \{va00191, s.fallah\}@surrey.ac.uk}}%
\thanks{$^{2}$I. Sharifi is with the Department of Mechanical and Aerospace Engineering, The George Washington University, Washington, DC
        {\tt\small{i.sharifi@gwu.edu}}}%
}

%
%

\markboth{Journal of \LaTeX\ Class Files,~Vol.~14, No.~8, August~2015}%
{Shell \MakeLowercase{\textit{et al.}}: Bare Demo of IEEEtran.cls for IEEE Journals}
%



\maketitle

\begin{abstract}

The problem with existing camera-based Deep Reinforcement Learning approaches is twofold: they rarely integrate high-level scene context into the feature representation, and they rely on rigid, fixed reward functions. To address these challenges, this paper proposes a novel pipeline that produces a neuro-symbolic feature representation that encompasses semantic, spatial, and shape information, as well as spatially boosted features of dynamic entities in the scene, with an emphasis on safety-critical road users. It also proposes a Soft First-Order Logic (SFOL) reward function that balances human values via a symbolic reasoning module. Here, semantic and spatial predicates are extracted from segmentation maps and applied to linguistic rules to obtain reward weights. Quantitative experiments in the CARLA simulation environment show that the proposed neuro-symbolic representation and SFOL reward function improved policy robustness and safety-related performance metrics compared to baseline representations and reward formulations across varying traffic densities and occlusion levels. The findings demonstrate that integrating holistic representations and soft reasoning into Reinforcement Learning can support more context-aware and value-aligned decision-making for autonomous driving.
\end{abstract}


%
\IEEEpeerreviewmaketitle

\section{Introduction}
\label{section:introduction}

Autonomous Driving (AD) is an ambitious goal that the research community and industry strive towards. To achieve this, the autonomous vehicle (AV) needs to navigate intricate urban environments, where ensuring human safety is paramount. Deep Reinforcement Learning (DRL) has emerged as a promising paradigm for learning driving policies directly from experiences. A trend in DRL research for AD is to rely on camera images as the primary input to the system, due to their low cost and rich semantic content.

However, learning driving policies from raw images poses significant challenges due to the high-dimensional input data and interpretability issues. To address this, researchers have explored region-based, spatial-temporal, and transformer-based attention mechanisms~\cite{Cultrera_2020_CVPR_Workshops}, \cite{cultrera2023explaining}, \cite{10736002}, \cite{https://doi.org/10.1049/itr2.12086}, \cite {10790926}, \cite{jia2025drivetransformerunifiedtransformerscalable}. A key limitation of these methods is that they primarily highlight the image regions the model focuses on. However, when humans drive, they rely on semantic understanding of objects in the scene and their spatial relations (e.g., a pedestrian is next to a crosswalk).

A possible approach to encoding this information is Hyperdimensional Computing (HDC) \cite{kanerva2009hyperdimensional}, which is a brain-inspired framework that encodes information as high-dimensional vectors (hypervectors). Prior work has applied HDC to represent features from single models~\cite{9892281}, \cite{10.1145/3526241.3530331}, \cite{Zou2022Memory}, \cite{10248004}, multiple models \cite{9892622}, or transformers \cite{10168629} in the hyperspace, demonstrating reduced memory and computation while improving performance and noise robustness. However, these approaches were limited to image classification and object recognition.  An alternative strategy, Vector Semantic Representation (VSR), was proposed by Neubert et al.~\cite{neubert2021vector}, which encodes semantic class, spatial location, and appearance information extracted from semantic segmentation maps into a unified hypervector representation. A key advantage of VSR is that the single scene descriptor preserves relational structure, while remaining compatible with gradient-based learning. Despite this property, VSR has only been used for visual place recognition, where its potential as a feature representation method for camera-based DRL in AD remains unexplored.

On top of feature representation, researchers have explored utilizing HDC for RL, primarily as a lightweight alternative to DNNs. Previous works have demonstrated how low-dimensional states~\cite{10.1145/3489517.3530668}, \cite{10.1145/3508352.3549437}, \cite{10.1145/3583781.3590298}, \cite{10.1145/3649476.3658760}, \cite{10.1145/3695875}, as well as LiDAR sensor data~\cite{10610176} can be represented as hypervectors and passed to a HD model, or an ensemble of HD models \cite{10.1145/3695875} for value-based (i.e., Q-Learning), and policy-based RL~\cite{10.1145/3489517.3530668}. Together these studies demonstrate the viability of HDC for control tasks. However, aside from~\cite{10610176}, these studies mainly focus on low-dimensional states or simplified environments. This raises questions on their suitability for complex tasks like AD, where the state-space can consists of high-dimensional sensor data.

Beyond representation, another fundamental challenge in DRL for AD lies in the design of the reward function, which plays a central role in shaping learned behaviors. For AVs to be reliably deployed in the real world, these behaviors must align with human values, which are ethical, cultural, and social principles that determine how AV systems make decisions, interact with humans, and function in society. Most prior works, such as~\cite{9582640}, \cite{10538211} use fixed reward functions that combine penalties and incentives into a weighted sum. While straightforward, this formulation fails in complex scenarios that require nuanced trade-offs between competing values that account for the driving scene's context.

To address these limitations, adaptive reward functions have been proposed that dynamically re-weight values based on observations, using cues from LiDAR \cite{8500441}, object detection \cite{8723530}, attention maps \cite{10611046}, or semantic risk assessments \cite{asodia2026offlinereinforcementlearningusing}. However, several adaptive reward functions switch between objectives using predefined conditions, enforcing a single dominant value at a given time. This style of hard switching can introduce brittle behavior in more complex situations, where more flexible decision-making is required.

A promising approach to overcoming this limitation is integrating symbolic reasoning into the reward function. Symbolic reasoning enables the explicit identification of key objects and relations in the environment and allow the application of implicit rules to derive new information that guides decision-making. Recent efforts, such as LogiCity \cite{NEURIPS2024_8196be81}, an abstract urban simulation platform with customizable First-Order Logic (FOL), reflect the growing interest in embedding symbolic reasoning into AV systems. Other studies have used symbolic modules to filter unsafe actions \cite{doi:10.1177/03611981251357006}, achieve more human-like driving via answer set programming~\cite{kothawade2021autodiscernautonomousdrivingusing}, \cite{Khan2025Neuro} and enhance interpretability~\cite{gilpin2021neuro}, \cite{luo2024endtoendneurosymbolicreinforcementlearning}. However, most existing approaches apply symbolic reasoning as a post hoc safety filter or architectural constraint, rather than leveraging it to shape the reward function itself. As a result, the potential of symbolic reasoning to directly influence value alignment during policy learning remains largely underexplored.

The contributions of this paper are as follows:
\begin{itemize}
    \item We extend VSR to encode both static and dynamic entities, with weighting schemes that emphasize vulnerable road users such as pedestrians.
    \item We fuse the symbolic hypervector with spatially attended feature embeddings to create a compact, context-rich scene descriptor for downstream decision-making.
    \item We propose a Soft First-Order Logic (SFOL) reward function that balances the human values of safety, efficiency, and comfort by reasoning over semantic and spatial predicates extracted from the scene.
\end{itemize}

The structure of the paper is as follows. Section \ref{section:prelim} introduces the core concepts and Section \ref{section:methodology}  provides key details on the VSR strategy and the SFOL reward function. Next, Section \ref{section:experiment_setup} explains the experiment scenario setup, training details, and comparative setups used for evaluation. Following this, Section \ref{section:results} presents the qualitative and quantitative results from the experimentation and Section \ref{section:discussion_future_work} discusses the results and proposes directions for future research. Finally, conclusions are drawn in Section \ref{section:conclusion}.

\section{Preliminaries}
\label{section:prelim}
This section provides an introduction to the theoretical background concepts of the proposed work, and includes: Reinforcement Learning, Hyperdimensional Computing, and Soft First-Order Logic.
\subsection{Reinforcement Learning (RL)}
RL control problems can be formulated as a Markov Decision Process (MDP), which is defined by the tuple $\{\mathcal{S}, \mathcal{A}, \mathcal{P}, \mathcal{R}, \gamma\}$. Here, at timestep $t$, the RL agent receives a state observation $s \in \mathcal{S}$ from the environment and performs an action $a \in \mathcal{A}$, according to its policy $\pi(a_t|s_t)$, to interact with the environment. Afterwards, the environment returns a numerical reward signal $r \in \mathcal{R}$ (discounted by the factor $\gamma$), which represents the value of taking action $a_t$ in state $s_t$ and the transition dynamics $\mathcal{P}: \mathcal{S} \times \mathcal{A} \rightarrow \mathcal{S}$, then takes the agent into the new state $s_{t+1}$. In episodic tasks, the agent remains in this interaction loop within the environment until it reaches a termination state $s_T$. The primary objective of the RL agent is to learn an optimal policy $\pi: \mathcal{S} \rightarrow \mathcal{A}$ that maximizes the expected cumulative discounted returns: $\text{max} \mathop{\mathbb{E}_\pi} [\sum_{t=1}^{T} \gamma ^tr_t]$. Specific details on how the vehicle longitudinal control task is formulated as an MDP are given in Section \ref{section:methodology_reinforcement_learning}. 
\subsection{Hyperdimensional Computing}
\label{section:prelim_hdc}
Within HDC scalar or symbolic data is encoded into high-dimensional binary or bipolar hypervectors. The dimensionality of these vectors is typically in the thousands (i.e., $1k\:\text{-}\:10k$ dimensions). A key aspect of the hypervectors is that they are constructed so that randomly sampled vectors are nearly orthogonal to each other with high probability \cite{kanerva2009hyperdimensional}, meaning that no two hypervectors are similar (i.e., their dot product is close to zero). This property allows distinct entities to be encoded without collisions or interference. In addition, by representing information across thousands of dimensions, if some dimensions are perturbed, the overall semantic meaning remains largely intact. Moreover, HDC relies on the following small set of algebraic operations to create structured and composite representations:
\begin{itemize}
    \item \textbf{Binding $\otimes$ :} Associates two pieces of information (i.e., key-value pairs), yielding a new hypervector dissimilar to its constituents.
    \item \textbf{Bundling $\oplus$ :} Aggregate multiple hypervectors into a single vector that preserves their similarities, enabling set-like representation. 
\end{itemize}
\noindent The operations explicitly encode role–filler pairs by binding abstract roles (e.g., spatial positions) to concrete fillers (e.g., semantic entities) within a fixed-dimensional space. This compositional approach enables complex symbolic structures to be represented as single hypervectors while preserving the ability to retrieve an approximation of the original components. In addition, multiple complex symbolic structures can be combined using the bundling $\oplus$ operator to produce a resultant hypervector descriptor. 

This paper uses \emph{Holographic Reduced Representations} (HRR) \cite{377968}, which is a real-valued instantiation of HDC that generates hypervectors by sampling a zero-mean Gaussian distribution. Moreover, in HRR, the binding operation $\otimes$ is implemented via circular convolution, and the bundling operation $\oplus$ is carried out by an element-wise addition operation. In limited prior work, HRR has been used within deep learning pipelines for multi-label classification \cite{NEURIPS2021_d71dd235} and recasting self-attention for malware detection \cite{pmlr-v202-alam23a}. These works highlight that HRR is well-suited for representing structured, relational information in continuous learning systems.
\subsection{Soft First-Order-Logic}
\label{section:prelim_soft_fol}
First-Order Logic (FOL) is a knowledge representation and reasoning (KRR) framework that provides a structured method to encode relationships between entities and to draw conclusions from a set of premises using logical inference. FOL enables systems to represent complex facts about the world and deduce new information through rule-based reasoning. Generally, rules in FOL are written as Horn clauses, which contain at most one positive literal, the head, and zero or more literals in the body. The rule syntax takes the following form:
\begin{equation}
    \texttt{H} \; \texttt{:-} \; \texttt{B}_1, \texttt{B}_2, ..., \texttt{B}_n\:,
\end{equation}
\noindent where, \texttt{H} is the rule head and $\texttt{B}_i$, $i=1$ denotes the atomic formulas that make up the rule body. If the body predicates $\texttt{B}_1$ to $\texttt{B}_n$ are true, then \texttt{H} is true; otherwise \texttt{H} is false.

Soft First-Order Logic (SFOL), as used in the context of Probabilistic Soft Logic (PSL), is an extension of classical FOL. Unlike classical FOL, which operates on binary truth values (true or false), SFOL allows variables to take on soft truth values within the interval $[0,1]$. These values represent varying degrees of truth, capturing the uncertainty inherent in many real-world scenarios.
PyReason \cite{aditya_pyreason_2023} is a neuro-symbolic reasoning engine designed to combine logical inference with uncertainty handling. Specifically, PyReason assigns each predicate, $P(x)$ an uncertainty bound $[\tau^-,\tau^+]$ to represents its degree of truth. The midpoint of the uncertainty bound can be used to represent the confidence value that the given predicate holds true, shown in:
\begin{equation}
    \tau(P(x)) \in [\tau^-,\tau^+]
    \label{eqn:uncertainty_bounds}\:, 
    \quad
    \hat\tau(P(x)) = \frac{\tau^-+\tau^+}{2}\:.
\end{equation}
During inference, PyReason uses these bounds to evaluate rules and propagate confidence levels through the reasoning process, enabling more flexible and context-aware decision-making.
\section{Methodology}
\label{section:methodology}
\begin{figure*}[!th]
\centering
\includegraphics[width = 1.0\linewidth]{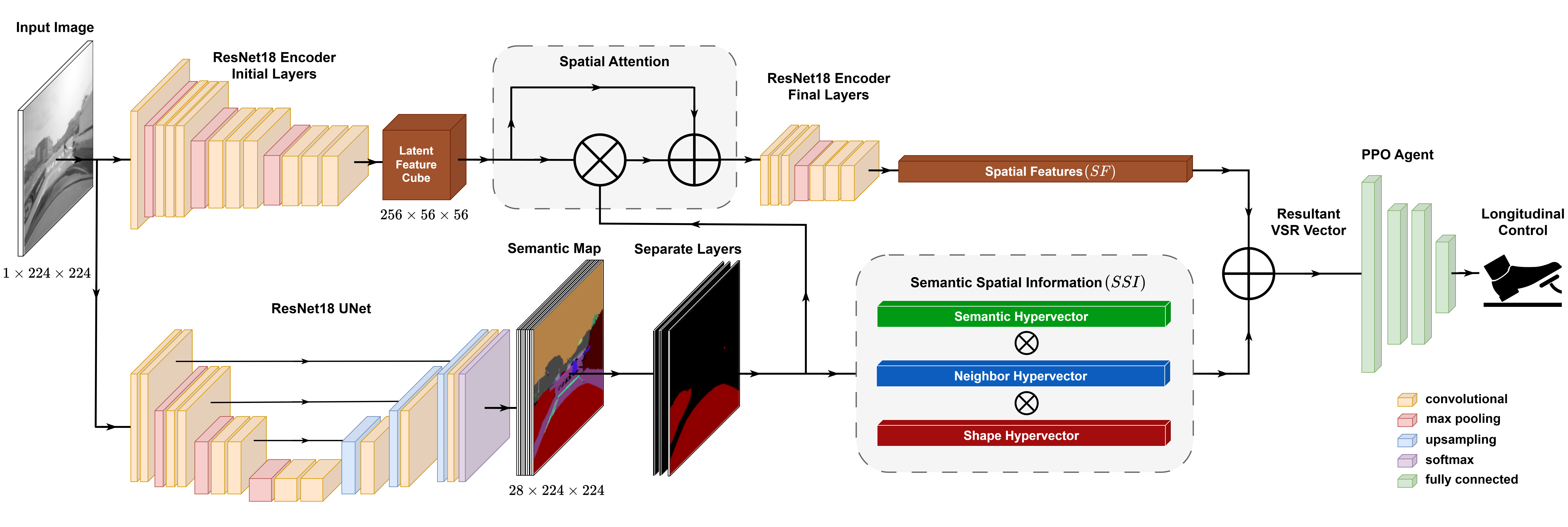}
\caption{Overview of the proposed pipeline. Grayscale camera images are passed through two parallel streams: a ResNet18 autoencoder that produces latent feature embeddings, and a ResNet18 UNet model to generate semantic maps. This map serves two functions: (1) to apply spatial attention to the latent features, and (2) to formulate a hypervector that encodes entity semantics and spatial relations. The resultant hypervector is combined with the spatial features to produce the final VSR vector, which is then fed into a PPO agent for longitudinal control.}
\label{fig:architecture}
\end{figure*}

This paper proposes a new HDC-based representation method to facilitate robust and sample-efficient RL training for AD. This section begins by presenting the construction of the VSR for driving scenes, which is illustrated in Figure \ref{fig:architecture}. Additionally, this paper proposes a novel SFOL reward function that integrates symbolic reasoning into the reward function. Key aspects of how the predicates are extracted and the rules under which they are applied to are detailed in this section. This process is illustrated in Figure \ref{fig:calculate_reward}.
\subsection{Semantic Segmentation Maps}
\label{section:methodology_semantic_segmentation_maps}
The components of the proposed VSR method rely on the analysis of semantic segmentation maps. Within the pipeline, semantic segmentation maps are generated by passing camera images through a UNet model \cite{10.1007/978-3-319-24574-4_28} with a ResNet18 backbone \cite{He_2016_CVPR} (shown in Figure \ref{fig:architecture}). The UNet is chosen for its ability to generate semantic maps in real-time inference speeds, which is essential for AV deployment. The resulting semantic map is a tensor of size $28\times224\times224$, where each channel represents the per-pixel probability of one of $28$ semantic classes.
\subsection{Vector Semantic Representation (VSR)}
\label{section:methodology_vsr}
The proposed VSR method aims to produce a holistic representation of the driving scene. To achieve this, each dynamic entity, $E_i$, is defined as:
\begin{equation}
    E_i = SSI_i \oplus SF_i\:,
    \label{eqn:entity_calculation}
\end{equation}
\noindent where $SSI_i$ represents the Spatial Semantic Information, and $SF_i$ denotes the Spatial Features. 
\subsubsection{Spatial Semantic Information ($SSI_i$)}
\begin{figure}[!th]
\centering
\includegraphics[width = 1.0\linewidth]{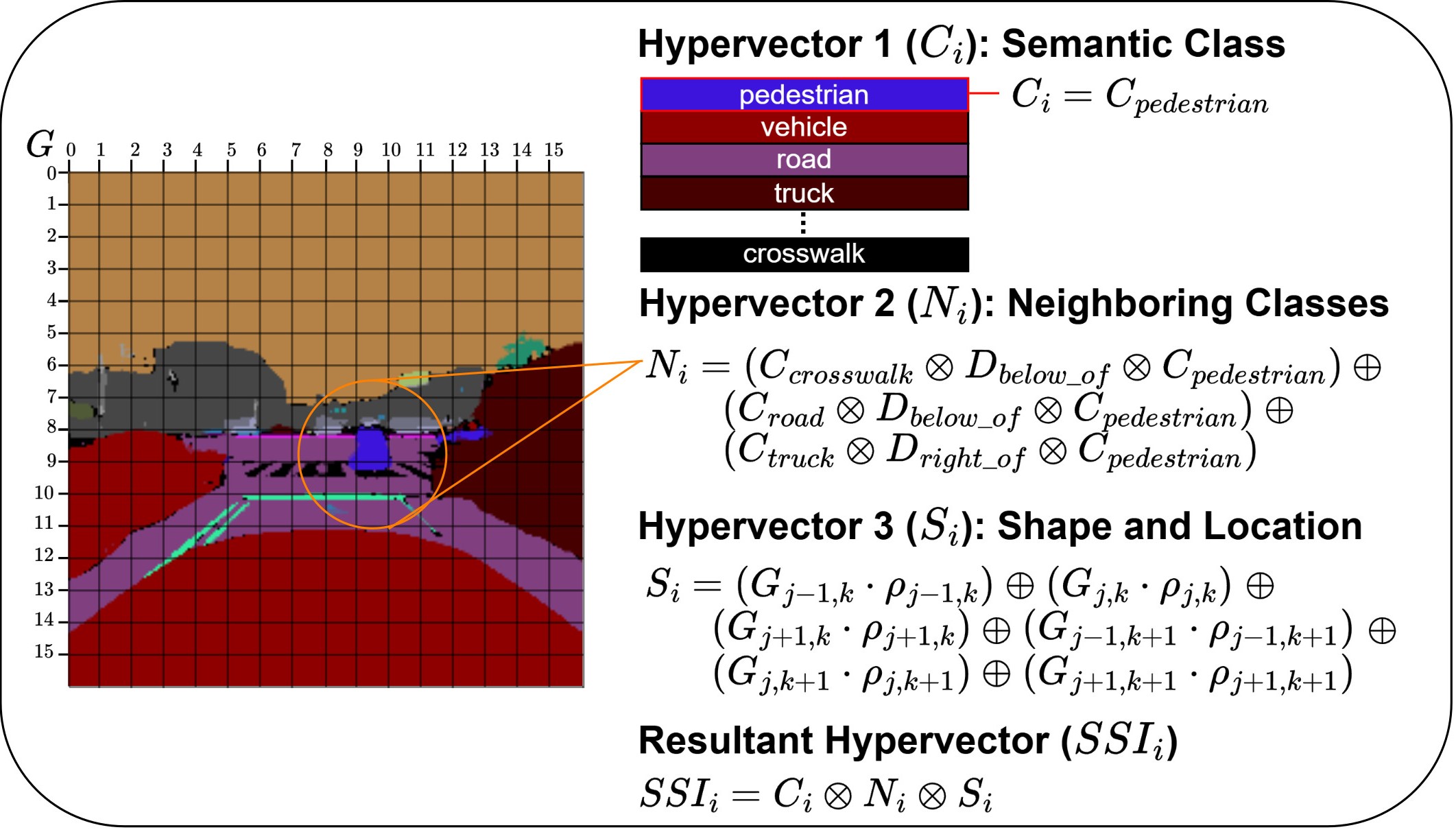}
\caption{An illustration of how the $SSI_i$ hypervector is formed for an entity $E_i$ (e.g., pedestrian).}
\label{fig:vsr}
\end{figure}
The $SSI_i$ for an entity represents $3$ key pieces of information: the semantic class the entity belongs to, the spatial relationships between the given entity and other semantic classes, as well as the shape and location of the entity in the semantic map. To note, all symbolic information is represented individually using HRR hypervectors, where each symbol $s \in \mathcal{S}$ is mapped to a unique hypervector $h_s \in \mathbb{R}^{2048}$. A dimensionality of $2048$ is chosen to match the $SF_i$ embeddings for later fusion. Figure \ref{fig:vsr} provides an example of how the $SSI_i$ is calculated for a pedestrian entity.

Beginning with the first hypervector, $C_i$, which encapsulates the entity's semantic class; a random hypervector is generated to represent each semantic class within the set $C_{class}:class \in \{pedestrian, vehicle, truck, road, crosswalk, sidewalk\}$. This subset of semantic classes is selected for its relevance to the occluded pedestrian crossing scenario and can be expanded to more classes as needed. As shown in Figure \ref{fig:vsr}, $C_1=C_{pedestrian}$, which encodes the information that the entity belongs to the pedestrian class.

The next hypervector is the neighbor hypervector, $N_i$, which represents the spatial information of neighboring semantic classes in relation to the given entity. For this stage, $3$ additional fixed random hypervectors are generated to represent the directional symbols $D_{direction}:direction \in \{left\_of, right\_of, below\_of\}$. Using the example from Figure \ref{fig:vsr}, to create a class-direction pairing, the binding ($\otimes$) operation is used to combine the neighbor class (e.g., $C_{crosswalk}$), the direction hypervector (e.g., $D_{below\_of}$), and the entity's semantic class(e.g., $C_{pedestrian}$), which represents a complex symbolic structure (e.g., "crosswalk below of pedestrian"). To identify each neighboring class, separate layers of the semantic map (within the set $C_{class}$) are taken, and a 15-cell window along the directional boundary of the given entity is searched for the appearance of any relevant classes. If multiple classes appear in various directions, the bundling ($\oplus$) operation is used to combine each class-direction hypervector, resulting in the final neighbor vector, $N_i$.

The final hypervector that makes up the $SSI_i$, is the shape and location hypervector, $S_i$. The proposed method determines the entity's location by first defining a $16\times16$ grid $G$ on the semantic layer to which the entity belongs (e.g., the pedestrian layer). To identify the grid cells occupied by the entity, a fixed random hypervector is generated for each grid cell, $G_{x,y}$. To estimate the entity's shape and location, the hypervectors of the grid cells that contain the given entity are combined using a weighted bundling operation. Each grid cell hypervector is weighted by the proportion of the cell's area occupied by the entity. Figure \ref{fig:vsr} illustrates an example of how the shape and location of a pedestrian entity are encoded as a single hypervector $S_i$. In the example, the pedestrian predominantly occupies grid cell $G_{j,k}$, with approximately $96\%$ of the cell covered; $j=9,k=8$ denotes the grid cell containing the entity's center point. This occupancy is reflected by the corresponding grid cell's hypervector $(G_{j,k} \cdot \rho_{j,k})$, where $\rho_{j,k}=0.96$ to represent the $96\%$ cell occupancy. The pedestrian also occupies the surrounding grid cells to a lesser extent, and is thus included in $S_i$ with proportionally smaller weights. It should be noted that in cases where multiple entities exist within a semantic class, the SciPy label function \cite{2020SciPy-NMeth} is used to identify separate entities. 

Finally, the $C_i, N_i, S_i$ hypervectors are combined with the binding operation to produce the single $SSI_i$ hypervector for the given entity as shown in:
\begin{equation}
    SSI_i = C_i \otimes N_i \otimes S_i\:.
    \label{eqn:ssi_calculation}
\end{equation}
\subsubsection{Spatial Features ($SF_i$)}
\label{section:methodology_semantic}
The $SF_i$ embeddings for an entity are produced using the spatial attention pipeline introduced in \cite{asodia2026offlinereinforcementlearningusing}. As shown in Figure \ref{fig:architecture}, images are encoded using a ResNet18-based autoencoder, where the encoder learns latent feature embeddings that capture the structural information of the input image.

Following \cite{10611046}, the entity-specific layers of the semantic map are used to apply spatial attention to the feature embeddings. This produces a set of spatially attended feature embeddings. The full details and design rationale for the spatial attention mechanism are provided in \cite{asodia2026offlinereinforcementlearningusing}.
\subsubsection{Final Vector Semantic Representation}
\label{section:method_final_vsr_vector}
To produce the final VSR vector, each dynamic entity $E_i$ is combined into a single vector using a weighted bundling operation, as shown in:
\begin{equation}
    VSR = \sum_{i=1}^k w_i \cdot E_i\:,
    \label{eqn:vsr_calculation}
\end{equation}
\noindent where $w_i$ is a weighting factor, dictated by the semantic class of the entity. The original authors \cite{neubert2021vector} determined $w_i$ based on the size of the $i$-th entity. However, this is counterintuitive for pedestrian avoidance, as the size of pedestrians is considerably smaller than that of vehicles. Instead, pedestrian entities are weighted by their perceived vulnerability, with pedestrians on the road or in the crosswalk assigned the highest weight. In contrast, pedestrians on the sidewalk are at less risk and thus receive a lower weight. Alternatively, the weightings for vehicle entities are tuned by their category (i.e., cars and trucks) to capture differences in occlusion levels rather than absolute collision risk. Specifically, larger vehicles are assigned a comparably higher weight than smaller vehicles, as they can occlude critical portions of the scene and reduce visibility of other road users, thus increasing uncertainty in downstream decision-making. The weighted bundling operation enables the representation to emphasize entities that require increased vigilance in safety-critical situations; the weights are listed in Table \ref{tbl:vsr_weights}. The final VSR vector is then passed downstream to the Proximal Policy Optimization (PPO) agent \cite{schulman2017proximalpolicyoptimizationalgorithms} to produce the vehicle's longitudinal control. 
\begin{table}[h]
\centering
\caption{Weighting factor, $w_i$ for VSR.}
\renewcommand{\arraystretch}{1.2} 
\begin{tabular}{c c}
\hline
\textbf{Dynamic Entity} & \textbf{Weight, $w_i$} \\
\hline
Pedestrian on Road/Crosswalk       & $1.0$ \\
Pedestrian on Sidewalk         & $0.50$ \\
Vehicle    & $0.50$ \\
Truck    & $0.75$ \\
\hline
\end{tabular}
\label{tbl:vsr_weights}
\end{table}
\subsection{Reinforcement Learning Setup}
\label{section:methodology_reinforcement_learning}
This subsection defines the vehicle longitudinal control task as an MDP and provides details on the state and action spaces. In addition, it will explain the base reward components, which encompass the key human values of safety, legality, efficiency, and comfort.
\subsubsection{State Space}
\label{section:methodology_state_space}
The state space consists of grayscale images (dimensions $1\times224\times224$) from a dashboard camera mounted on the ego vehicle. 
\subsubsection{Action Space}
\label{section:methodology_action_space}
The PPO agent will output a single continuous action $a_t$, between the range $[-1,1]$, representing the ego vehicle's throttle and brake values at each timestep $t$. 
\subsubsection{Base Reward Components}
The proposed method for integrating an SFOL reasoning module into the reward function is applied to the base reward components from \cite{8500441}. These components are selected because they reflect key human values commonly associated with vehicle longitudinal control. The first component represents the value of safety and is defined as:
\begin{equation}
    g_{saf}(x_{t}, u_{t}) = -(\zeta \frac{v_{t}^2}{d_{t}+\epsilon} + \eta \textbf{1}(d_{t} = 0))\:,
    \label{eqn:safe}
\end{equation}
\noindent where $\zeta>0$ represents a weight to penalize the ego vehicle for traveling at a high speed. The variable $v_t$ denotes the ego vehicle’s velocity, and $d_t$ corresponds to the distance between the vehicle and the pedestrian. To offset the denominator, a small constant $\epsilon$ is introduced. In addition, an indicator function $\textbf{1}(d_{t} = 0)$ is included that will apply a penalty $\eta>0$ when the condition $d_t=0$ holds true (i.e., when the ego vehicle and the pedestrian collide). If the condition is false, then no additional penalty is applied.

The second component covers the value of efficiency and is defined in Equation \ref{eqn:efficient}. Moreover, the component provides a reward proportional to the ego vehicle's speed (weighted by $\lambda>0$), incentivizing the ego vehicle to drive at faster speeds.  
\begin{equation}
    g_{eff}(x_{t}, u_{t}) = \lambda v_{t}\:,
    \label{eqn:efficient}
\end{equation}
The final component ensures the encapsulation of the smoothness value by penalizing the ego vehicle for large decelerations via the weight $\xi>0$. This component decreases the jerkiness of the ego vehicle's longitudinal control. Below is the definition of the final component:
\begin{equation}
    g_{smooth}(x_{t}, u_{t}) = -\xi(v_{t} - v_{t+1})^2 = -\xi(a_{t}\Delta t)^2\:, 
    \label{eqn:smooth}
\end{equation}
The final reward signal is the summation of the base reward components, weighted by an SFOL reasoning module. The next subsection provides key details on how this reasoning module is configured.
\subsection{Soft First-Order-Logic Reward Function}
\label{section:methodology_reward_function}
\begin{figure}[!th]
\centering
\includegraphics[width = 0.99\linewidth]{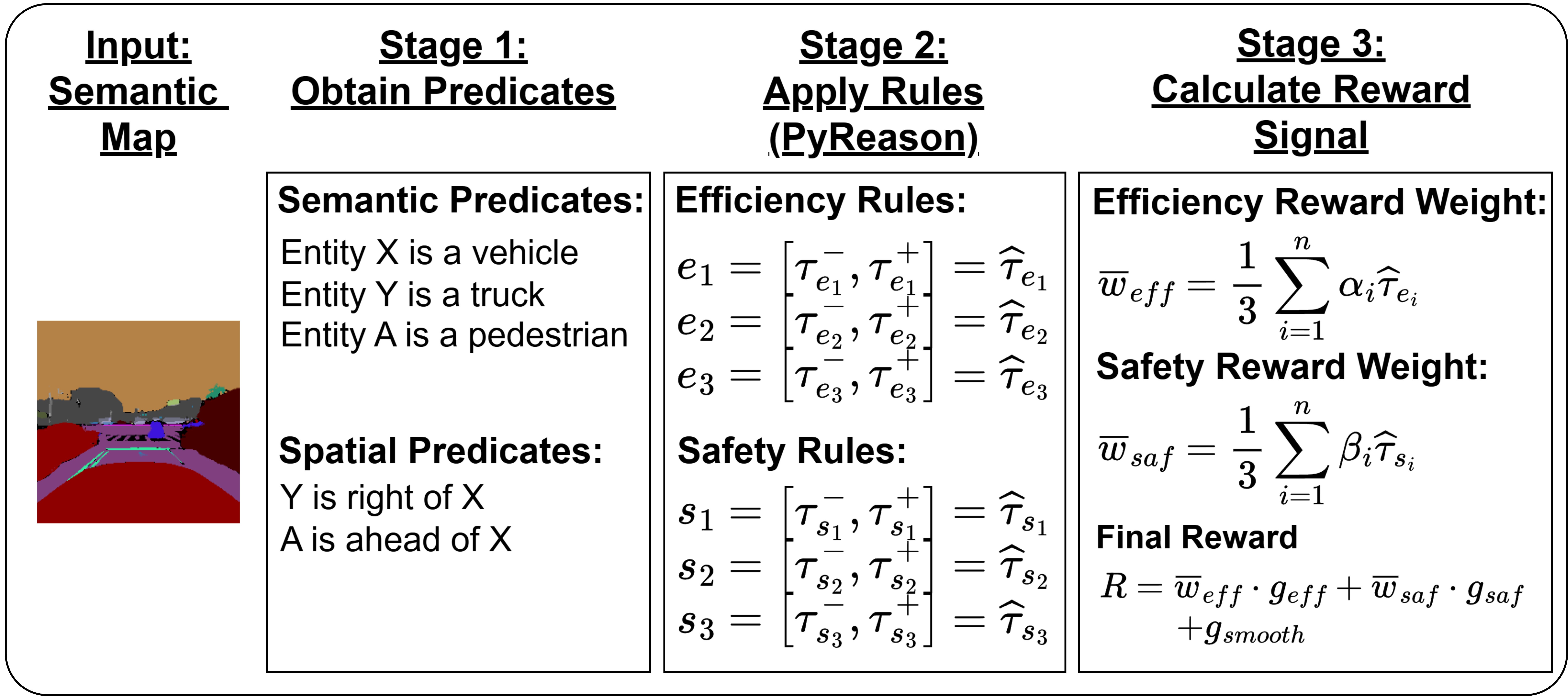}
\caption{Illustration of the SFOL reward computation. Predicates extracted from the semantic map are applied to linguistic rules, producing rule confidence values that determine the safety and efficiency reward weights in Equation \ref{eqn:final_reward}.}
\label{fig:calculate_reward}
\end{figure}
Figure \ref{fig:calculate_reward} illustrates the stages to obtain the reward signal from the SFOL reward function (implemented using PyReason). This subsection explains each stage in detail, starting with the method for obtaining predicates from the driving scene, followed by a breakdown of the linguistic rules. Finally, the manner in which the reward weights are calculated and how they are used in the final reward calculation is covered.
\subsubsection{Obtain Predicates}
To begin, predicates are extracted by analyzing the relevant layers of the semantic segmentation maps. The set of predicates is outlined in Table \ref{tbl:predicates}, and they are composed of three types: semantic, spatial, and a special type to represent explicit negation (details provided later). 
\begin{table}[h]
\centering
\caption{Set of Predicates obtained from the driving scene.}
\renewcommand{\arraystretch}{1.2} 
\begin{tabular}{p{0.15\textwidth} p{0.04\textwidth} >{\raggedright\arraybackslash}p{0.22\textwidth}}
\hline
\textbf{Predicates} & \textbf{Arity} & \textbf{Description} \\
\hline
\texttt{Vehicle(X)}       & 1 & Entity X is a vehicle. \\
\texttt{Truck(X)}         & 1 & Entity X is a truck. \\
\texttt{Pedestrian(X)}    & 1 & Entity X is a pedestrian. \\
\texttt{Crosswalk(X)}      & 1 & Entity X is a crosswalk. \\
\hline
\texttt{LeftOf(X,Y)}      & 2 & Entity X is left of entity Y. \\
\texttt{RightOf(X,Y)}     & 2 & Entity X is right of entity Y. \\
\texttt{CenterOf(X,Y)}    & 2 & Entity X is center of entity Y. \\
\texttt{IsNear(X,Y)}      & 2 & Entity X is near entity Y. \\
\texttt{Approaching(X,Y)} & 2 & Entity X is approaching entity Y. \\
\texttt{OnRoad(X)}        & 1 & Entity X is on the road. \\
\texttt{OnCrosswalk(X)}    & 1 & Entity X is on the crosswalk. \\
\texttt{Occludes(X,Y)}    & 2 & Entity X occludes entity Y. \\
\hline
\texttt{IsClear(X)}       & 1 & Entity X is clear of occlusions. \\
\texttt{NoCrosswalk(X)}    & 1 & \makecell[l]{Entity X is on a road with \\no crosswalk.} \\
\texttt{EmptyLane(X)}     & 1 & Entity X is in an empty lane. \\
\hline
\end{tabular}
\label{tbl:predicates}
\end{table}
Semantic predicates encode entity membership in semantic classes (e.g., entity X is a pedestrian) and are extracted from the semantic layers. Specifically, for a given layer (e.g., pedestrian layer), the SciPy label function is used to identify clusters of pixels for a given class, where each cluster $c$ is treated as a known entity with confidence $\hat\tau$ given by the mean softmax value, as shown in: 
\begin{equation}
    \hat\tau_c = \frac{1}{|R_c|}\sum_{p\in R_c}\sigma(p)\:,
    \label{eqn:entity_confidence_calcuation}
\end{equation}
\noindent where $|R_c|$ denotes the number of pixels in cluster $c$ and $\sigma(p)$ represents the softmax probability of pixel $p$ belonging to a given class. To obtain the predicate's uncertainty bounds $\tau$, $5\%$ of both sides of the confidence value $\hat\tau$ is used $\tau(P(x))=[0.95\hat\tau, 1.05\hat\tau]$. To sufficiently build the knowledge base for reasoning, entities within the pedestrian, vehicle, truck, and crosswalk layers, along with their uncertainty bounds, are recorded.  

The next type extracted is spatial predicates, which represent the spatial relationship between two semantic predicates. Spatial predicates are binary and encode ego-relative spatial relationships (left, center, and right). To identify these spatial relationships relative to the ego vehicle, the semantic map is divided into three equal vertical regions. The region where the center point of a dynamic entity lies determines the spatial predicate asserted. For example, if an entity is in the rightmost region of the camera image, then the predicate "RightOf(entity, ego)" can be asserted. In addition, a fundamental computer vision technique, the pinhole method \cite{10.5555/1941882}, is used to infer the distance, $d_{cross}$, between the ego vehicle and a detected crosswalk. The ego vehicle can be considered as approaching the crosswalk if $50m \leq d_{crosswalk} \leq 200m$, and if $d_{crosswalk} < 50m$, the ego vehicle is classed as "at the crosswalk". The pinhole method was used over other methods like radar, as it enables specific targeting of objects in the image observation. 

Furthermore, spatial predicates relative to each pedestrian entity are taken into account, as their location is crucial information for safe AV control. For each pedestrian, a $15$ cell window below and beside the entity is analyzed to determine if they are standing on the sidewalk, road, or crosswalk, or if they are next to any vehicles. Finally, the predicate "Occludes(X,Y)" is defined to represent the fact that "entity X occludes entity Y". This predicate is extracted by interpreting the spatial ordering of semantic predicates from the ego-centric view. An entity is considered occluded if another entity lies closer to the ego vehicle along the same line of sight. Altogether, the spatial predicates offer additional information that is integral for AD reasoning.

Finally, PyReason employs negation-as-failure: if the system cannot prove a predicate to be true, it assumes it is false. This differs from explicit negation, in which a predicate must be stated as false. Thus, to represent explicit negation for calculating the reward weights, a special set of predicates is defined. For example, in the safety component, a key spatial predicate is whether a detected crosswalk is occluded. Equally important for the efficiency component, however, is the opposite condition: verifying that the crosswalk is clear. If a crosswalk contains no pedestrians, the predicate IsClear(A) is asserted. Beyond crosswalks, special predicates are also needed to capture the general driving condition (e.g., when there are no crosswalks or vehicles in the lane). In these instances, the predicates NoCrosswalk(X) and EmptyLane(X) are asserted.
\subsubsection{Rule Base}
\begin{table*}[h]
\centering
\caption{Set of Safety and Efficiency Rules for the SFOL reward function.}
\renewcommand{\arraystretch}{1} 
\begin{tabular}{>{\raggedright\arraybackslash}p{0.28\textwidth} >{\raggedright\arraybackslash}p{0.68\textwidth}}
\hline
\textbf{Rule} & \textbf{Description} \\
\hline
\texttt{\makecell[l]{\textbf{efficiency1(X)} :- Vehicle(X), \\NoCrosswalk(X), EmptyLane(X).}} & 
If X is a vehicle and X is on a road that has no crosswalk and no other vehicles in its current lane, then prioritize efficiency. \\
\texttt{\makecell[l]{\textbf{efficiency2(X)} :- Vehicle(X), \\Crosswalk(A), approaching(X,A), \\IsClear(A).}} & 
If X is a vehicle, and A is a crosswalk, and X is approaching A, and A is clear, then prioritize efficiency. \\
\texttt{\makecell[l]{\textbf{efficiency3(X)} :- Vehicle(X), \\Crosswalk(A), IsAt(X,A), \\IsClear(A).}} & 
If X is a vehicle and A is a crosswalk, and X is at A and A is clear, then prioritize efficiency. \\
\texttt{\makecell[l]{\textbf{safe1(X)} :- Vehicle(X), \\Pedestrian(P), OnRoad(P), \\CenterOf(P,X).}} & 
If X is a vehicle and P is a pedestrian, and P is on the road, and P is center of X, then prioritize safety. \\
\texttt{\makecell[l]{\textbf{safe2(X)} :- Vehicle(X), \\Pedestrian(P), OnCrosswalk(P).}} & 
If X is a vehicle and P is a pedestrian, and P is on the crosswalk, then prioritize safety. \\
\texttt{\makecell[l]{\textbf{safe3(X)} :- Vehicle(X), \\Vehicle(Y), Crosswalk(A), \\RightOf(Y,X), Occludes(Y,A).}} & 
If X is a vehicle and Y is a vehicle and A is a crosswalk, and Y is right of X and Y occludes A, then prioritize safety. \\
\texttt{\makecell[l]{\textbf{safe3(X)} :- Vehicle(X), \\Vehicle(Y), Crosswalk(A), \\CenterOf(Y,X), Occludes(Y,A).}} & 
If X is a vehicle and Y is a vehicle and A is a crosswalk, and Y is center of X and Y occludes A, then prioritize safety. \\
\texttt{\makecell[l]{\textbf{safe3(X)} :- Vehicle(X), \\Vehicle(Y), Crosswalk(A), \\LeftOf(Y,X), Occludes(Y,A).}} & 
If X is a vehicle and Y is a vehicle and A is a crosswalk, and Y is left of X and Y occludes A, then prioritize safety. \\
\hline
\end{tabular}
\label{tbl:rule_set}
\end{table*}
Once the predicates are extracted, they are applied to a rule base using PyReason as the interface engine (Stage $2$ of Figure \ref{fig:calculate_reward}). Table \ref{tbl:rule_set} lists the safety and efficiency rules in Horn clause form. Each reward component consists of $3$ rules composed with the body predicates from Table \ref{tbl:predicates}. For prioritizing efficiency, the first rule covers the most likely case, in which there are no crosswalks or vehicles ahead of the ego vehicle. The next two efficiency rules cover more subtle instances when the ego vehicle is either at or approaching a crosswalk that is clear of occlusions. 

For the safety rules, "safe1" covers the most safety-critical scenario: a pedestrian directly in front of the ego vehicle. The next rule encompasses instances where a pedestrian is on a crosswalk in view of the ego vehicle. The final rule addresses the most subtle context, where the crosswalk is occluded by entities to the left, right, or in front of the ego vehicle. The motivation for including this rule is to introduce some caution into decision-making. To represent logical disjunction (i.e., OR operation) in PyReason, multiple rules with the same head ("safe3") are established to represent the occlusion directions. If any one of the rule bodies for "safe3" is satisfied, the head predicate holds true.

At each timestep, PyReason performs an inference run over the extracted predicates. The average lower and upper bounds of the body predicates are used to represent the uncertainty bounds of the head predicate $H(x)$ (shown in Equation \ref{eqn:head_bounds}). To represent the confidence value that a given rule holds true, the midpoint of the head predicate's uncertainty bound is used.
\begin{equation}
\tau_H^- = \frac{1}{n} \sum_{i=1}^n \tau_i^-\:, 
\quad 
\tau_H^+ = \frac{1}{n} \sum_{i=1}^n \tau_i^+\:, 
\quad 
\hat{\tau}_H = \frac{\tau_H^- + \tau_H^+}{2}\:.
    \label{eqn:head_bounds}
\end{equation}
When multiple rules share the same head ("safe3"), PyReason aggregates uncertainty by taking the maximum lower and upper bounds. This is consistent with logical disjunction, where each rule independently supports the conclusion.
\subsubsection{Calculate Reward Signal}
The confidence values of the efficiency and safety rules are used to compute reward weights, as shown in Equation \ref{eqn:reward_weight}:
\begin{equation}
    \bar{w}_{eff} = \frac{1}{3} \displaystyle\sum_{i=1}^n \alpha_i \hat{\tau}_{e_i}\:,
    \quad
    \bar{w}_{saf} = \frac{1}{3} \displaystyle\sum_{i=1}^n \beta_i \hat{\tau}_{s_i}\:,
    \label{eqn:reward_weight}
\end{equation}
\noindent where $\alpha_i$ and $\beta_i$ determine the relative importance of each efficiency $\hat{\tau}_{e_i}$ and safety $\hat{\tau}_{s_i}$ rule in the final reward weighting. A weighted average is used because some rules apply only to edge cases and should not outweigh more critical indicators. For both the efficiency and safety rule set, two depend on the presence of a crosswalk, while one does not. 
Once the reward weights are determined, they can be used to weigh the aforementioned safety (Equation \ref{eqn:safe}) and efficiency (Equation \ref{eqn:efficient}) base reward components. The final reward calculation is shown in:
\begin{equation}
    R_{final} = \bar{w}_{saf} \cdot g_{saf} + \bar{w}_{eff} \cdot g_{eff} + g_{smooth}\:.
    \label{eqn:final_reward}
\end{equation}
\section{Experimental Setup}
\label{section:experiment_setup}
This section details the experimentation procedure and comparative setups used to evaluate the proposed pipeline. In addition, a definition of the occluded pedestrian crossing scenario set up in the CARLA Urban Driving Simulator \cite{Dosovitskiy17} is supplied. CARLA provides diverse, dynamic actors and sensor configurations that enable the setup of customizable edge-case scenarios.
\subsection{Occluded Pedestrian Crossing Scenario using CARLA}
\label{section:scenario}
\begin{figure}[!th]
      \centering
      \setlength{\fboxsep}{0pt}%
      \setlength{\fboxrule}{0pt}%
      \framebox{\parbox{3in}{\includegraphics[width=\linewidth]{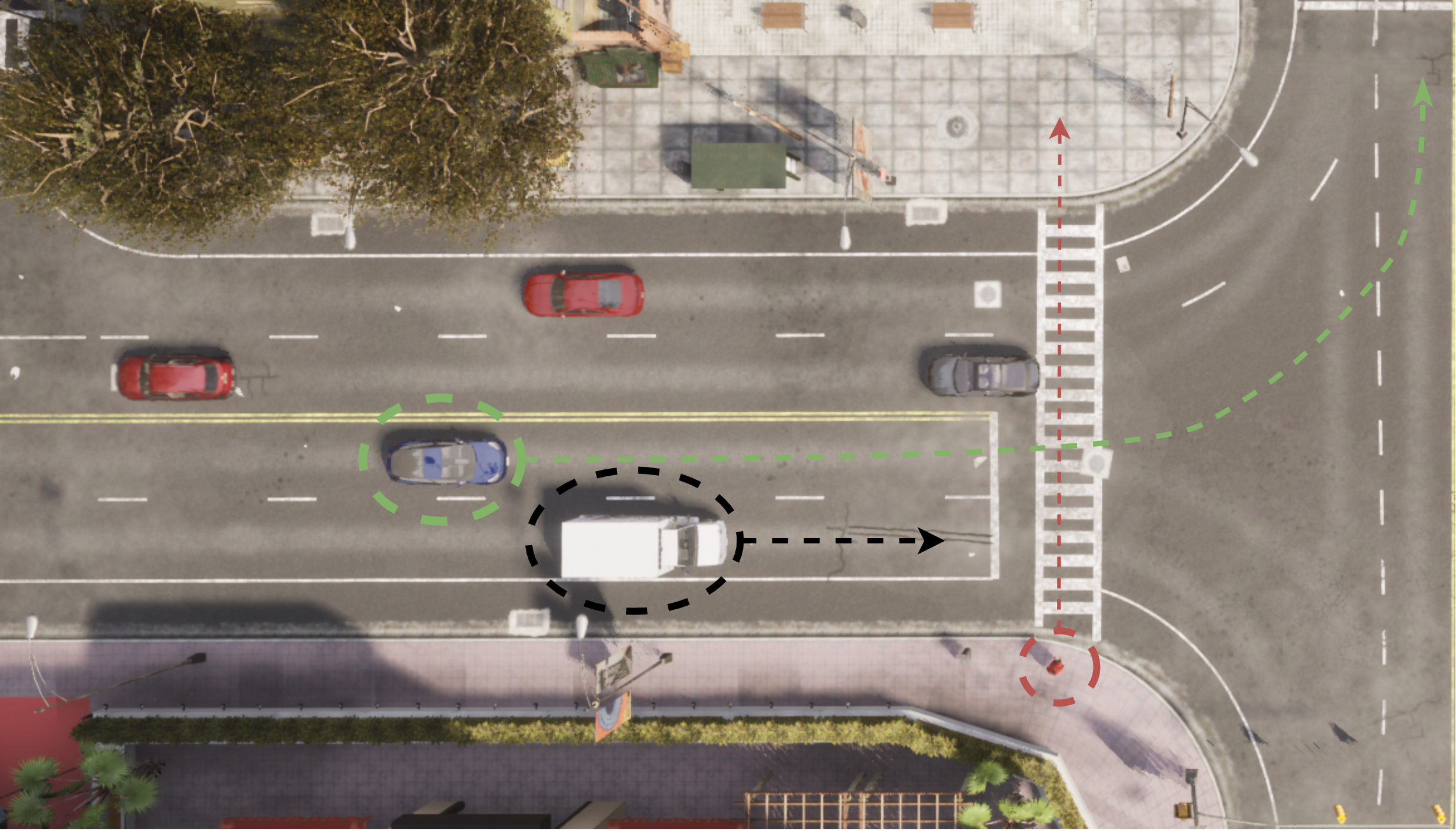}}

}
      \caption{Illustration of the occluded pedestrian crossing scenario in CARLA. The ego vehicle (circled in Green) starts at one end of the road and must navigate toward the goal whilst another vehicle (circled in Black) occludes part of the crosswalk. As it approaches, a pedestrian (circled in Red) enters the crosswalk, requiring the ego vehicle to yield before proceeding safely.}
      \label{fig:carla-eb}
   \end{figure}
The occluded pedestrian crossing scenario models a safety-critical urban driving situation, where the ego vehicle must reach a goal point along a road that contains a crosswalk occluded by a stationary vehicle. Moreover, there are $2$ pedestrian behavior configurations: $1)$ the pedestrian is briefly visible before becoming occluded by the stationary vehicle, then enters the crosswalk as the ego vehicle approaches, and $2)$ the pedestrian is fully occluded until they suddenly cross the road. In addition, different occlusion levels are included, with a large van fully occluding the pedestrian and a smaller vehicle partially occluding the pedestrian. The agent’s goal is to output the vehicle's longitudinal control to safely reach the goal point, whilst avoiding collisions. Figure \ref{fig:carla-eb} illustrates the scenario within CARLA.

The pipeline’s robustness is evaluated with the following densities: "Low Ped Traffic" with $10$ pedestrians, "Med Ped Traffic" with $25$ pedestrians ($20\%$ running at $6km/h$ and $20\%$ crossing unexpectedly), and "High Ped Traffic" with $50$ pedestrians ($40\%$ running at $6km/h$ and $40\%$ crossing unexpectedly).

The ego vehicle is equipped with an RGB dashboard camera and a collision sensor that registers any collisions between the ego vehicle and other objects.
\subsection{Comparative Setups}
\label{section:comparative_setup}
There are two sets of comparative setups that are used to evaluate different components of the proposed pipeline.

The first set includes different integration strategies for the VSR vector. The motivation for these setups is to identify the optimal methods for fusing the $SSI$ hypervector with the $SF$ embeddings, and to assess how each component contributes to the learned policy. To note, the following setups utilize the proposed SFOL reward function:
\begin{itemize}
    \item $SSI \oplus SF$: element-wise addition is used to combine the $SSI$ hypervector and the $SF$ embeddings.
    \item $SSI \otimes SF$: element-wise multiplication is used to combine the $SSI$ hypervector and the $SF$ embeddings.
    \item $SSI$ Only: only the $SSI$ hypervector is passed downstream to the PPO agent.
    \item $SF$ Only: only the $SF$ embedding is passed downstream to the PPO agent, feature extraction approach from \cite{asodia2026offlinereinforcementlearningusing}.
\end{itemize}
The second set introduces variations of the SFOL reward function, where the rules outlined in Table \ref{tbl:rule_set} are removed. The rationale behind these setups is to evaluate the necessity of each rule. For each component (safety and efficiency), there is a rule that covers an explicit situation (i.e., rules "efficiency1" and "safe1") and rules that cover more subtle, rarer situations (i.e., rules "efficiency2", "efficiency3", "safe2", and "safe3"). If the complete rule set for a component is used, this is denoted as "CE" or "CS". However, if the subtle rules are removed, the resulting rule set is considered a partial rule set and is denoted as "PE" or "PS". The SFOL reward function variations are as follows; "CE \& CS", "CE \& PS", "PE \& CS" and "PE \& PS". Also, the risk factor adaptive reward function from \cite{asodia2026offlinereinforcementlearningusing} is included within the set to compare how the reward components are balanced.
\subsection{Model Evaluation} 
\label{section:model_evaluation}
The model evaluation consists of a series of quantitative and qualitative experiments. First, the training performance curves for the different VSR integration strategies are collected. This data provides an insight into the effectiveness of each strategy as a feature representation for policy learning. 

Next, a collision test is carried out on the VSR integration strategies in the occluded pedestrian crossing scenario under the different traffic densities outlined in Section \ref{section:scenario}. More specifically, for each setup under each traffic density and occlusion level, the success, collision, and stall rates, as well as the average stopping distance over $100$ evaluation episodes, are recorded. This experiment enables observation of how varying congestion and occlusion levels affect policy decision-making. In addition to the VSR integration strategies, a collision test is conducted across variations of the SFOL reward functions. 

In parallel with the collision tests, additional driving characteristic metrics are collected across the $100$ evaluation episodes. These metrics serve as an additional breakdown of the quality of vehicle control produced by the learned policies. The total set of driving characteristic metrics are as follows:
\begin{itemize}
    \item RMS(Speed): root mean square of vehicle speed.
    \item RMS(Acceleration)*: root mean square of vehicle acceleration.
    \item RMS(Jerk)*: root mean square of vehicle jerk.
    \item Min TTC: minimum time to collision.
    \item Impact Speed*: average vehicle speed at point of collision.
    \item Near Miss*: rate of near misses, which is defined as TTC $\leq 2s$.
    \item False Brake*: rate of false positive braking.
    \item TTG*: time to goal.
\end{itemize}
\noindent These metrics are normalized, and the metrics with an asterisk (*) are inverted ($1-metric$) to ensure consistency, so higher values correlate with better performance. For readability, these metrics can be categorized as safety metrics (Min TTC, Impact Speed, Near Miss, and False Brake), efficiency metrics (TTG), and comfort metrics (RMS(Speed), RMS(Acceleration), and RMS(Jerk)).

Finally, for quality assessment, the speed profile over a trajectory is recorded along with the weightings of each rule in the proposed SFOL reward function. This information enables the visualization of the smoothness of vehicle control and the contribution of individual rules to the final safety and efficiency reward weightings.
\section{Results} 
\label{section:results}
\subsection{VSR Integration Comparison} 
\label{section:results_hdc_comparison}
\subsubsection{Training Performance} 
\label{section:results_hdc_comparison_training_performance}
\begin{figure}[!h]
    \centering
    \includegraphics[width=1.0\linewidth]{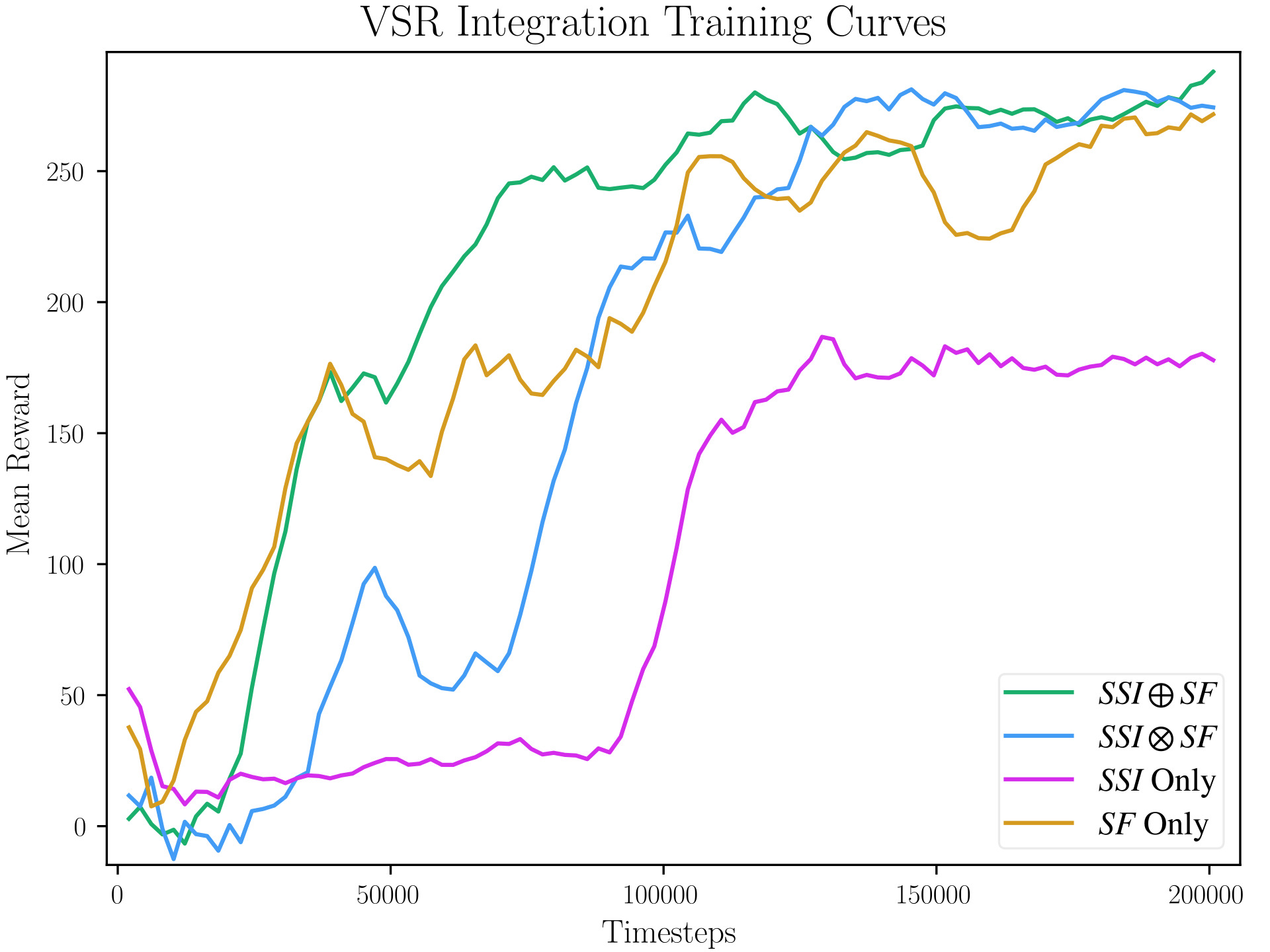}
    \caption{Mean reward over training timesteps for different VSR integration strategies: via element-wise addition ($SSI \oplus SF$ in Green), and via element-wise multiplication ($SSI \otimes SF$, in Blue), as well as using $SSI$ hypervector only ($SSI$ Only, in Magenta), and using the $SF$ embeddings only ($SF$ Only, in Yellow).}
    \label{fig:training-performance}
\end{figure}
Figure \ref{fig:training-performance} presents the training curves for different strategies for integrating the $SSI$ hypervector with the $SF$ embeddings using the proposed SFOL reward function. The $SSI$ Only setup exhibited the slowest convergence and plateaued the lowest mean reward ($\approx170$), indicating that the $SSI$ hypervector alone does not provide sufficient information for effective policy learning. Conversely, the $SF$ Only setup showed the fastest initial convergence, but exhibited slower and less stable improvements thereafter. Both the $SF$ Only and $SSI \otimes SF$ strategies achieved similar final rewards ($\approx265$). However, the $SSI \otimes SF$ setup converged more slowly while showing greater stability in the latter stages of training. The $SSI \oplus SF$ setup offered the most promising results, demonstrating fast initial convergence, stable learning, and yielding the highest mean reward ($\approx270$). These findings highlight that the $SSI$ hypervector alone is insufficient for robust policy learning. However, integrating the hypervector with the spatial features via element-wise addition, provides a more efficient and stable learning signal to the policy.
\subsubsection{Collision Test} 
\label{section:results_hdc_comparison_collision_test}
\begin{table*}[!th]
\centering
\caption{Collision test results from running each setup for 100 episodes in the partially occluded pedestrian crossing scenario. Each cell reports Partial (P) and Full (F) occlusion results side-by-side. S = Successful, C = Collision, T = Timeout, SD = Stopping Distance.}
\label{tbl:collision_test_vsr}
\resizebox{\textwidth}{!}{%
\begin{tabular}{c cccccccc cccccccc cccccccc}
\toprule
& 
\multicolumn{8}{c}{\textbf{Low Traffic Density}} &
\multicolumn{8}{c}{\textbf{Medium Traffic Density}} &
\multicolumn{8}{c}{\textbf{High Traffic Density}} \\
\cmidrule(lr){2-9} \cmidrule(lr){10-17} \cmidrule(lr){18-25}
& \multicolumn{2}{c}{\textbf{$\uparrow$ S (\%)}} & \multicolumn{2}{c}{\textbf{$\downarrow$ C (\%)}} & \multicolumn{2}{c}{\textbf{$\downarrow$ T (\%)}} & \multicolumn{2}{c}{\textbf{$\uparrow$ SD (m)}} 
& \multicolumn{2}{c}{\textbf{$\uparrow$ S (\%)}} & \multicolumn{2}{c}{\textbf{$\downarrow$ C (\%)}} & \multicolumn{2}{c}{\textbf{$\downarrow$ T (\%)}} & \multicolumn{2}{c}{\textbf{$\uparrow$ SD (m)}} 
& \multicolumn{2}{c}{\textbf{$\uparrow$ S (\%)}} & \multicolumn{2}{c}{\textbf{$\downarrow$ C (\%)}} & \multicolumn{2}{c}{\textbf{$\downarrow$ T (\%)}} & \multicolumn{2}{c}{\textbf{$\uparrow$ SD (m)}} \\
\cmidrule(lr){2-9} \cmidrule(lr){10-17} \cmidrule(lr){18-25}
\textbf{Setups} & P & F & P & F & P & F & P & F 
& P & F & P & F & P & F & P & F 
& P & F & P & F & P & F & P & F \\
\midrule
$SF$ only           & 98 & 96 & 2  & 4  & 0 & 0 & 3.24 & 3.51  & \textbf{97} & 94 & \textbf{3}  & 6  & 0 & 0 & 3.17 & 3.29  & 86 & 88 & 14 & 12 & 0 & 0 & 3.27 & 3.40 \\
$SSI \oplus SF$          & \textbf{99} & \textbf{99} & \textbf{1}  & \textbf{1}  & 0 & 0 & 3.15 & 4.15  & \textbf{97} & 95 & \textbf{3}  & 5  & 0 & 0 & 3.67 & 4.06  & \textbf{97} & \textbf{95} & \textbf{3}  & \textbf{5}  & 0 & 0 & 3.99 & 4.11 \\
$SSI \otimes SF$          & 98 & 98 & 2  & 2  & 0 & 0 & \textbf{5.00} & \textbf{5.66}  & \textbf{97} & \textbf{97} & \textbf{3}  & \textbf{3}  & 0 & 0 & \textbf{4.91} & \textbf{5.79}  & 96 & \textbf{95} & 4  & \textbf{5}  & 0 & 0 & \textbf{5.19} & \textbf{5.88} \\
$SSI$ only                & 87 & 87 & 13 & 13 & 0 & 0 & 2.05 & 2.06  & 76 & 77 & 24 & 23 & 0 & 0 & 2.07 & 2.19  & 70 & 67 & 30 & 33 & 0 & 0 & 2.18 & 2.06 \\
\bottomrule
\end{tabular}
}
\end{table*}
Table \ref{tbl:collision_test_vsr} summarizes the collision test results across traffic densities and occlusion levels. There is a consistent downward trend in success rates and an increase in collision rates as the traffic density rises, reflecting the inherent challenges of navigating congested environments. The performance of each setup tended to remain the same or slightly decrease under full occlusion compared to partial occlusion, as expected due to the agent's reduced visibility.

The $SF$ Only configuration performs reliably under low and medium traffic densities, but degrades substantially in high-density scenarios, where collision rates notably increase. This indicates that in congested environments, visual features alone may not be sufficient for robust decision-making.

Moreover, the $SSI \oplus SF$ setup achieved the most balanced performance, maintaining high success rates, low collision rates, and longer average stopping distances across the traffic densities and occlusion levels. Notably, this setup remained stable across both occlusion levels as traffic density increased, demonstrating greater robustness than $SF$ Only. 

Interestingly, the $SSI \otimes SF$ setup performed comparably to $SSI \oplus SF$ in success, collision, and stall rate, even surpassing it under the medium traffic density variant with full occlusion. However, this comes at the cost of substantially longer stopping distances, indicating excessively conservative behavior that may lead to inefficiencies and traffic flow problems if deployed in the real world.

In contrast, the $SSI$ Only setup achieved the worst overall performance, demonstrating the shortest average stopping distances, that lead to the highest collision rates. This highlights the importance of the $SF$ embeddings for producing a meaningful representation of the image scene.

Finally, the $SSI \oplus SF$ setup offered the most robust and balanced performance across varying traffic densities and occlusion levels.
\subsubsection{Driving Characteristic Metrics} 
\label{section:results_driving_characteristic_metrics}
\begin{figure}[t]
    \centering
    \begin{subfigure}[t]{0.49\columnwidth}
        \centering
        \includegraphics[width=\linewidth]{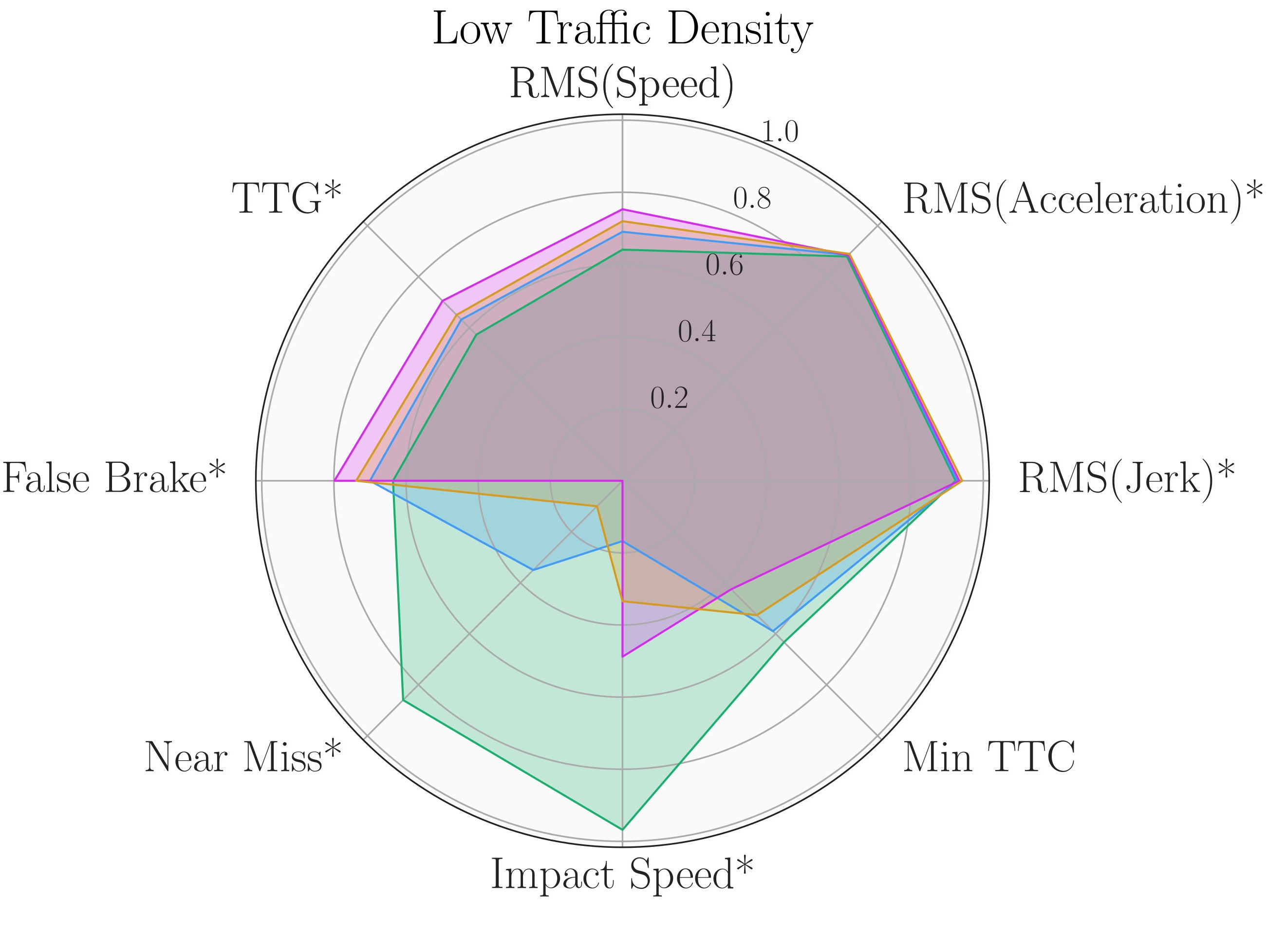}
        \caption{Low traffic density}
        \label{fig:low}
    \end{subfigure}
    \hfill
    \begin{subfigure}[t]{0.49\columnwidth}
        \centering
        \includegraphics[width=\linewidth]{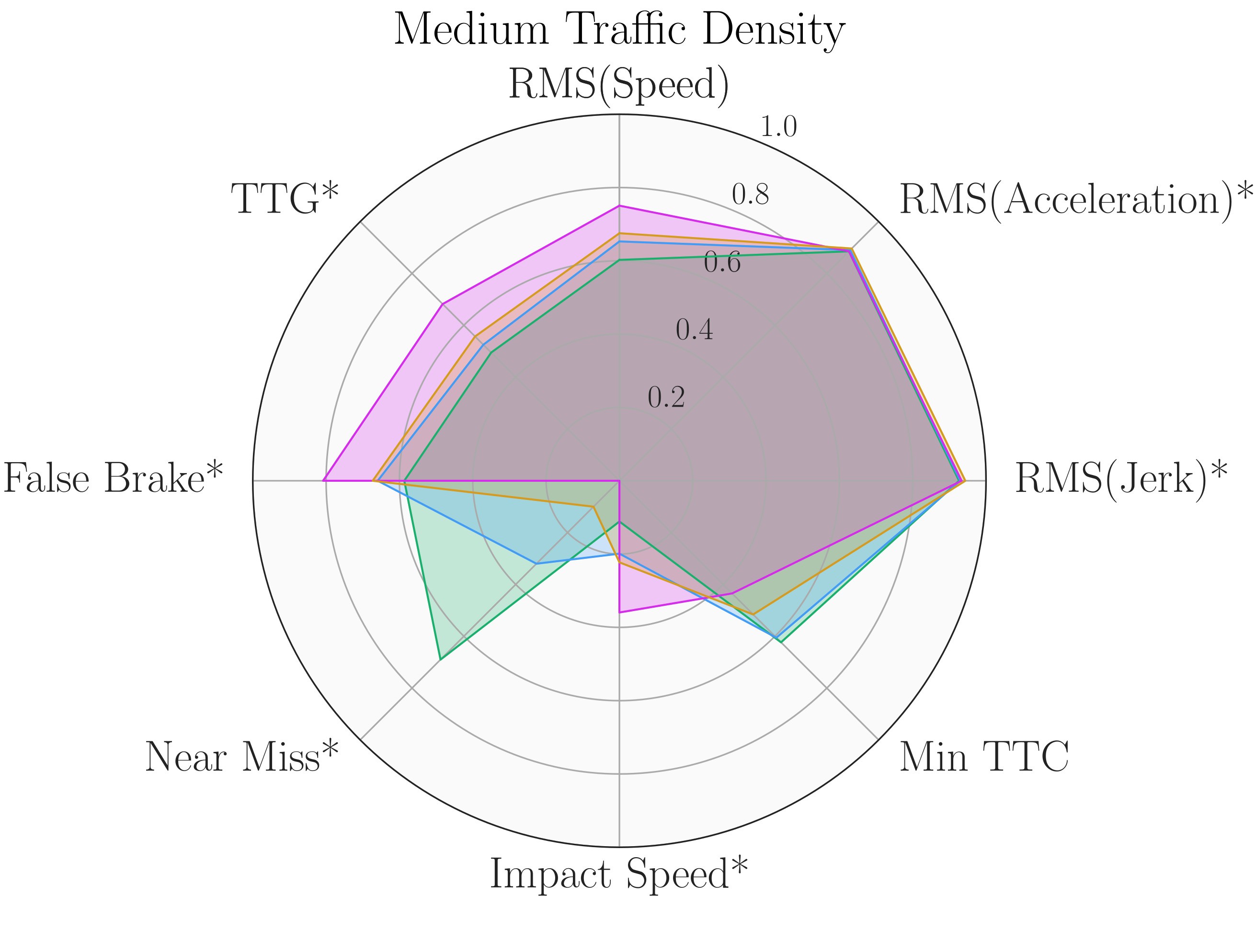}
        \caption{Medium traffic density}
        \label{fig:med}
    \end{subfigure}
    \begin{subfigure}[t]{1.0\columnwidth}
        \centering
        \includegraphics[width=\linewidth]{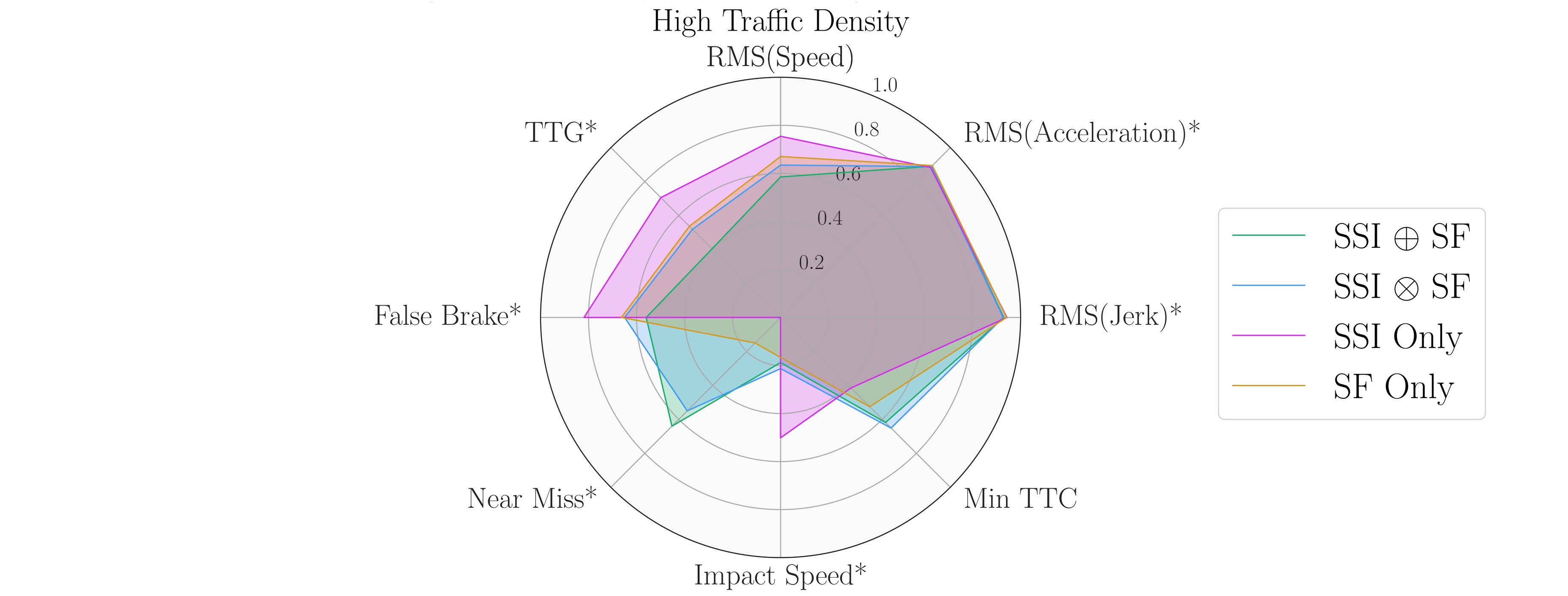}
        \caption{High traffic density}
        \label{fig:high}
    \end{subfigure}

    \caption{Driving characteristic metrics for traffic densities.}
    \label{fig:radar_metrics}
\end{figure}
Figure \ref{fig:radar_metrics} shows $3$ radar plots that present the driving characteristic metrics across low, medium, and high traffic densities. Across all settings, the comfort metrics remain largely similar between VSR integration strategies, suggesting that feature representation has minimal effect on driving smoothness.

The $SSI$ Only strategy reliably displayed the lowest minimum TTC and highest near-miss rates, reflecting an aggressive policy. This is supported by the strategy regularly attaining the highest RMS speed and TTG, and maintaining a moderate impact speed. These results reinforce the collision test findings in that the $SSI$ hypervector alone does not capture the explicit information needed for safe, reactive driving.

The $SF$ Only strategy demonstrated slightly improved near-miss rates and minimum TTC values compared to $SSI$ Only, but with a lower TTG and a higher false-brake rate, suggesting a more conservative yet reactive policy. This strategy's safety metrics noticeably degrade as the traffic density increases, indicating that relying on image features alone leads to delayed reactions to crossing pedestrians. 

Conversely, both $SSI \oplus SF$ and $SSF \otimes SF$ demonstrate meaningful improvements in the safety metrics across traffic densities. Initially, under low traffic density, the differences between the two strategies are most evident in near miss rates and impact speeds. However, the gap in safety metrics narrows as congestion increases, with both strategies converging on similar near-miss rates, impact speeds, and minimum TTCs. These improvements came with an increase in TTG, reflecting more cautious driving behavior.

Overall, the findings support the conclusions from the collision test (Table \ref{tbl:collision_test_vsr}). Neither symbolic nor spatial representations alone were sufficient for safe longitudinal control in heavily congested settings. Instead, using explicit spatial information alongside a symbolic representation that captures high-level context can yield a robust policy that performs well in challenging settings. 
\subsubsection{Trajectory Analysis}
\begin{figure}[!h]
    \centering
    \includegraphics[width=1.0\linewidth]{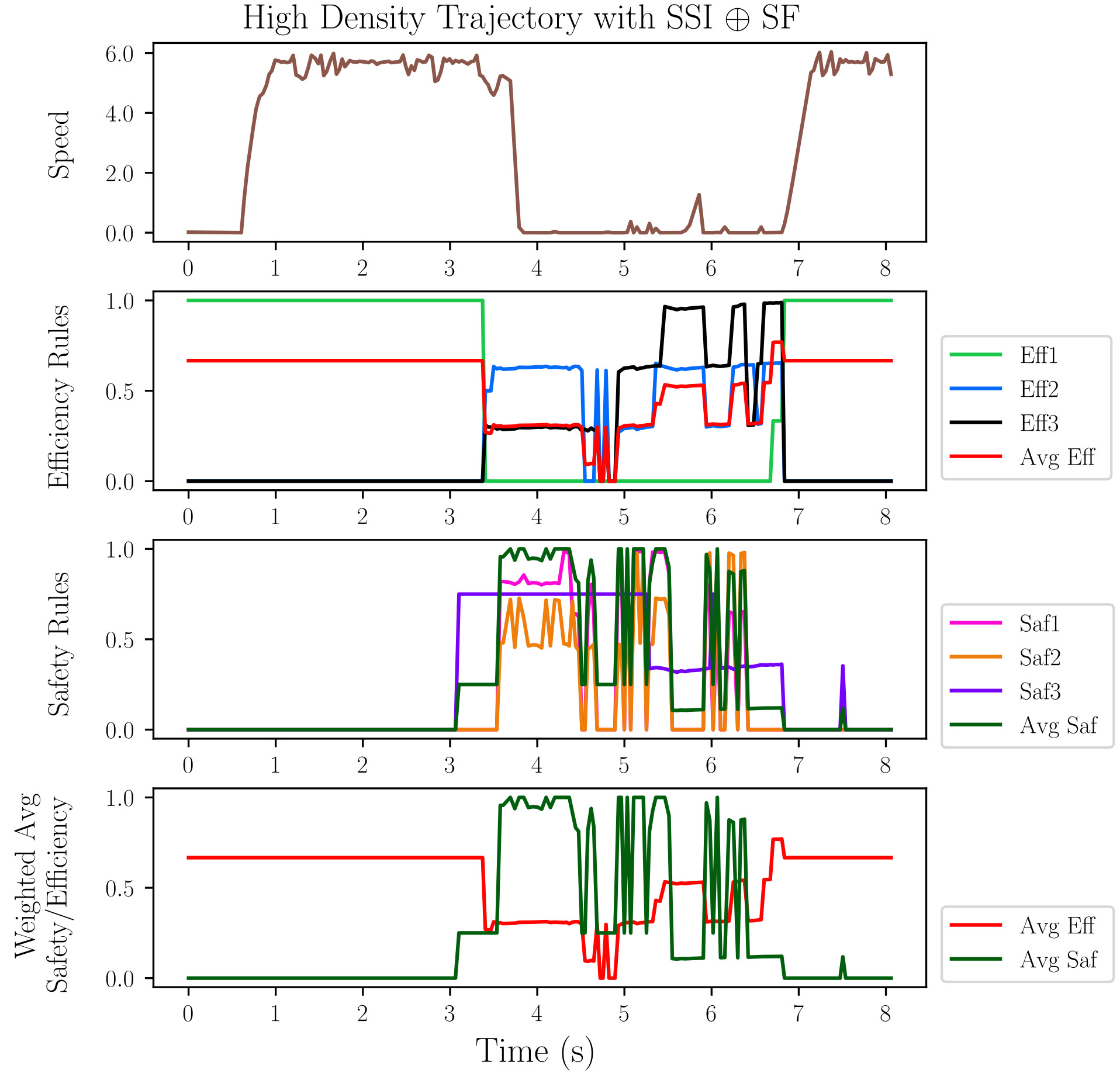}
    \caption{Row of plots illustrating the speed profile and activation of safety and efficiency rules. The bottom plot shows the average safety and efficiency weights used in the reward function.}
    \label{fig:rule_firings}
\end{figure}
This subsection presents a qualitative assessment of the $SSI \oplus SF$ strategy under a high traffic density, which was selected due to its performance in the quantitative performance in earlier experiments. Figure \ref{fig:rule_firings} shows the speed profile alongside the confidence values for each efficiency and safety rule and the corresponding reward weights. Given that the reward function is not used during inference, the reward weights are included for post-hoc analysis.

The top graph illustrates the speed profile, where the ego vehicle sharply accelerates from a stationary position to a desirable speed, approaches a crosswalk, and decelerates to a stop as pedestrians begin to cross. The ego vehicle remains stationary whilst the crosswalk is occupied, and takes off once the pedestrians clear the scene. Generally, the vehicle control was relatively stable, apart from a brief false takeoff, whilst the crosswalk was occupied, indicating an instance of misalignment between perception and decision-making.

Moving to the rule firings, the second graph illustrates the confidence values of the efficiency rules. Initially, the "efficiency1" rule attains its maximum confidence, and other rules remain low, as the ego vehicle is in an empty lane and has not yet reached a crosswalk. When the ego vehicle approaches the crosswalk, confidence in "efficiency1" falls, whilst the confidence in "efficiency2" and "efficiency3" increases, reducing the average efficiency weight. As the ego vehicle transitions from approaching to being at an occupied crosswalk, these two rules alternate in dominance. Notably, "efficiency3" exhibits sharp oscillations during this period, indicating the event of pedestrians entering and leaving the crosswalk. Once the crosswalk clears, confidence in "efficiency1" rises back to a high level as the ego vehicle accelerates toward the goal point. 

The third graph describes the trajectory from the safety rules' perspective. At first, all safety rules stay at zero, indicating the road is clear, with no crosswalks. As the ego vehicle approaches a crosswalk occluded by surrounding vehicles, confidence in "safe3" increases. Soon after, "safe1" and "safe2" quickly rise as pedestrians enter the crosswalk. The weighted average safety value then increases, shifting priority to safety and causing the vehicle to stop at the crosswalk. During this period, "safe1" and "safe2" fluctuate as pedestrians enter and exit, while "safe3" decreases, reflecting fewer occlusions by other vehicles. Once the crosswalk is clear, confidence in all safety rules falls, allowing the vehicle to accelerate toward its goal. 

Finally, the bottom graph shows the weighted average of safety and efficiency confidence values, used to balance the SFOL reward function. In most cases, the system favors the appropriate human value for each context; however, near the crosswalk, the system shows pronounced fluctuations that directly affect the stability of the reward weights. Although the system is quick to recover by reasserting safety, the brief instability exposes a potential vulnerability in the reasoning process, which may affect control smoothness in real-time deployment.
\subsection{SFOL Reward Function Ablation Study} 
\label{section:results_reward_ablation}
\subsubsection{Collision Test}
\label{section:results_reward_ablation_collision_test}
Table \ref{tbl:collision_test_rule_ablation} reports the collision test results for different SFOL reward function variants using the $SSI \oplus SF$ integration strategy. The results for $SSI \oplus SF$ presented in Table \ref{tbl:collision_test_vsr} are included in this table as the CE \& CS variation of the SFOL reward function.

The Risk Factors baseline yielded high success rates in low and medium traffic densities, with acceptable average stopping distances. However, the brittleness of this style of reasoning becomes apparent in high traffic density, with success rates decreasing in partial- and fully-occluded settings, respectively. In contrast, the CE \& CS configuration, which integrates the complete efficiency and safety rule sets, delivers the strongest overall performance with the least performance decrease. 

Removing the subtle safety indicators (CE \& PS) leads to very short average stopping distances and signals aggressive driving and delayed reaction to pedestrians. Conversely, removing the subtle efficiency indicators (PE \& CS) results in overly conservative behavior, with the longest stopping distances and the highest timeout rates in dense traffic. While focusing on safety led to better performance than the CE \& PS variation, being overly conservative can still be dangerous in real-world AD.

Finally, the PE \& PS configuration, which only retains the explicit rules, yields moderate performance in low-density scenarios but degrades as traffic density increases. The exhibited behavior is similar to the Risk Factors approach, but lacks the flexibility needed for complex interactions, highlighting the limitations of a reduced reasoning capacity. 
\begin{table*}[!h]
\centering
\caption{Collision test results from running each setup for 100 episodes in the occluded pedestrian crossing scenario, showing both Partial (P) and Full (F) occlusion. S = Successful, C = Collision, T = Timeout, SD = Stopping Distance.}
\label{tbl:collision_test_rule_ablation}
\resizebox{\textwidth}{!}{%
\begin{tabular}{c cccccccc cccccccc cccccccc}
\toprule
& 
\multicolumn{8}{c}{\textbf{Low Traffic Density}} &
\multicolumn{8}{c}{\textbf{Medium Traffic Density}} &
\multicolumn{8}{c}{\textbf{High Traffic Density}} \\
\cmidrule(lr){2-9} \cmidrule(lr){10-17} \cmidrule(lr){18-25}
& \multicolumn{2}{c}{\textbf{$\uparrow$ S (\%)}} & \multicolumn{2}{c}{\textbf{$\downarrow$ C (\%)}} & \multicolumn{2}{c}{\textbf{$\downarrow$ T (\%)}} & \multicolumn{2}{c}{\textbf{$\uparrow$ SD (m)}} 
& \multicolumn{2}{c}{\textbf{$\uparrow$ S (\%)}} & \multicolumn{2}{c}{\textbf{$\downarrow$ C (\%)}} & \multicolumn{2}{c}{\textbf{$\downarrow$ T (\%)}} & \multicolumn{2}{c}{\textbf{$\uparrow$ SD (m)}} 
& \multicolumn{2}{c}{\textbf{$\uparrow$ S (\%)}} & \multicolumn{2}{c}{\textbf{$\downarrow$ C (\%)}} & \multicolumn{2}{c}{\textbf{$\downarrow$ T (\%)}} & \multicolumn{2}{c}{\textbf{$\uparrow$ SD (m)}} \\
\cmidrule(lr){2-9} \cmidrule(lr){10-17} \cmidrule(lr){18-25}
\textbf{Setups} & P & F & P & F & P & F & P & F 
& P & F & P & F & P & F & P & F 
& P & F & P & F & P & F & P & F \\
\midrule
Risk Factors           & 98 & 97 & 2  & 3  & 0 & 0 & 3.83 & 4.01  & 95 & 95 & 5  & 5  & 0 & 0 & 3.96 & 4.12  & 91 & 89 & 9  & 11 & 0 & 0 & 3.89 & 4.21 \\
CE \& CS         & \textbf{99} & \textbf{99} & \textbf{1}  & \textbf{1}  & 0 & 0 & 3.15 & 4.15  & \textbf{97} & 95 & \textbf{3}  & 5  & 0 & 0 & 3.67 & 4.06  & \textbf{97} & \textbf{95} & \textbf{3} & \textbf{5} & 0 & 0 & 3.99 & 4.11 \\
CE \& PS               & 9  & 11 & 91 & 89 & 0 & 0 & 0.85 & 1.04  & 3  & 5  & 97 & 95 & 0 & 0 & 1.17 & 1.27  & 2  & 3  & 98 & 97 & 0 & 0 & 1.28 & 1.06 \\
PE \& CS               & 73 & 91 & 8  & 2  & 19 & 7 & \textbf{5.94} & \textbf{6.14} & 85 & \textbf{97} & 3  & \textbf{3} & 12 & 0 & \textbf{5.89} & \textbf{6.02} & 91 & 79 & 6  & 4  & 3 & 27 & \textbf{5.99} & \textbf{6.36} \\
PE \& PS               & 94 & 91 & 6  & 9  & 0 & 0 & 3.51 & 4.12  & 88 & 89 & 12 & 10 & 0 & 1 & 3.84 & 3.93  & 79 & 88 & 21 & 12 & 0 & 0 & 3.21 & 2.06 \\
\bottomrule
\end{tabular}
}
\end{table*}
\section{Discussion \& Future Work}
\label{section:discussion_future_work}
Across the VSR integration strategies, the quantitative and qualitative results indicate that element-wise addition offers the most effective balance of training efficiency and robustness in congested environments. Whilst the training curves showed that $SF$ Only converged quickly, the collision test highlighted that relying on SF alone leads to sub-optimal performance as congestion increases. Throughout the experiments, $SSI$ Only consistently underperformed, demonstrating that the $SSI$ hypervector alone is insufficient for effective policy learning. The integration of both components has shown to improve performance in complex settings, however, using element-wise multiplication may be too aggressive compared to element-wise addition, which enables a more stable fusion by minimizing destructive interference between representations. It is important to note that no variations of the pipeline was able to achieve a $100\%$, leaving room for additional refinements. Future work could incorporate temporal symbolic information into the hypervector using the permutation HDC operation, which could reduce fluctuations in the decision-making.

The SFOL reward function ablation study highlights the value of incorporating symbolic reasoning to balance components of the reward function over using rigid risk-based approaches, enabling more adaptive decision-making. However, the rule firings for the trajectory using the complete SFOL reward function exposes erroneous fluctuations in heavily congested settings, suggesting that the reasoning module needs to be improved. Future work could extend the symbolic module with traffic law knowledge, incorporating abductive reasoning to handle hidden hazards, and introduce counterfactual reasoning to learn from near-miss scenarios.
\section{Conclusion} 
\label{section:conclusion}
In conclusion, this paper presented a pipeline that integrated a hypervector encapsulating the semantic and spatial information of dynamic objects, along with their spatially attended feature embeddings, to produce a single VSR vector. Entities within this VSR vector were weighted to prioritize vulnerable road users. In addition, this paper proposed a Soft First-Order Logic (SFOL) reward function that enabled more human-centered autonomous emergency braking by balancing safety, efficiency, and comfort through symbolic reasoning. A PPO agent was used for longitudinal control, and the pipeline was validated across different traffic densities in the occluded pedestrian crossing scenario set up in CARLA. The results indicated that integrating the two representation components via element-wise addition and utilizing a comprehensive rule set for reward reasoning achieved the most robust performance, maintaining high success rates and low collision rates under high traffic congestion. 


%





\ifCLASSOPTIONcaptionsoff
  \newpage
\fi



\bibliographystyle{IEEEtran}
\bibliography{main}

%



%
\begin{IEEEbiography}[{\includegraphics[width=1in,height=1.25in,clip,keepaspectratio]{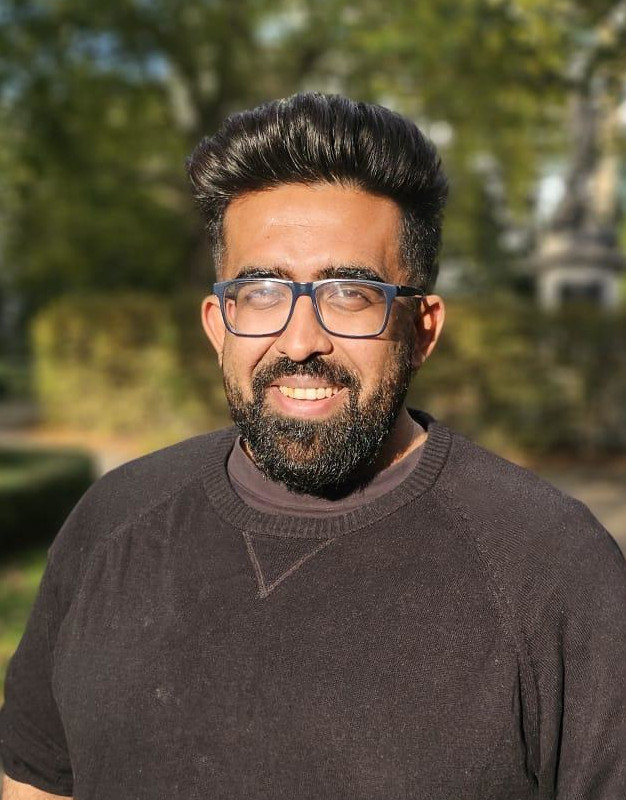}}]{Vinal Asodia}
is a PhD Researcher in the Connected and Autonomous Vehicles (CAV) Lab at the University of Surrey. He received his B.Sc. degree in Computer Science from the University of Surrey. His research focuses on developing end-to-end deep reinforcement learning frameworks for autonomous driving systems that are explainable and aligned with human values. His research interests include enhancing the transparency, safety, and social acceptability of AI-driven autonomous systems, contributing to the reliability of these technologies in real-world applications. He has published his work in the IEEE International Conference on Robotics and Automation (ICRA).
\end{IEEEbiography}
\begin{IEEEbiography}[{\includegraphics[width=1in,height=1.25in,clip,keepaspectratio]{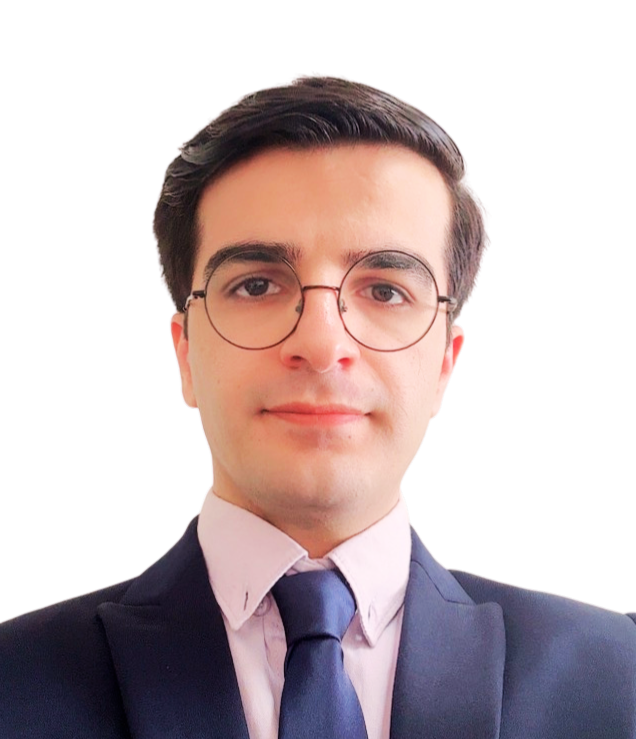}}]{Iman Sharifi} is a Ph.D. student in Mechanical and Aerospace Engineering at George Washington University. He received his M.Sc. in Mechanical Engineering from Sharif University of Technology and his B.Sc. in Aerospace Engineering from K. N. Toosi University of Technology. His research focuses on neuro-symbolic artificial intelligence, multi-agent reinforcement learning, and safety-critical autonomous systems, with applications to unmanned aerial systems, advanced air mobility, and autonomous driving. He has conducted research funded by NASA’s University Leadership Initiative (ULI). His work has appeared in venues such as IJCAI, Transportation Research Record, Applied Sciences, and Smart Agricultural Technology.
\end{IEEEbiography}
\begin{IEEEbiography}[{\includegraphics[width=1in,height=1.25in,clip,keepaspectratio]{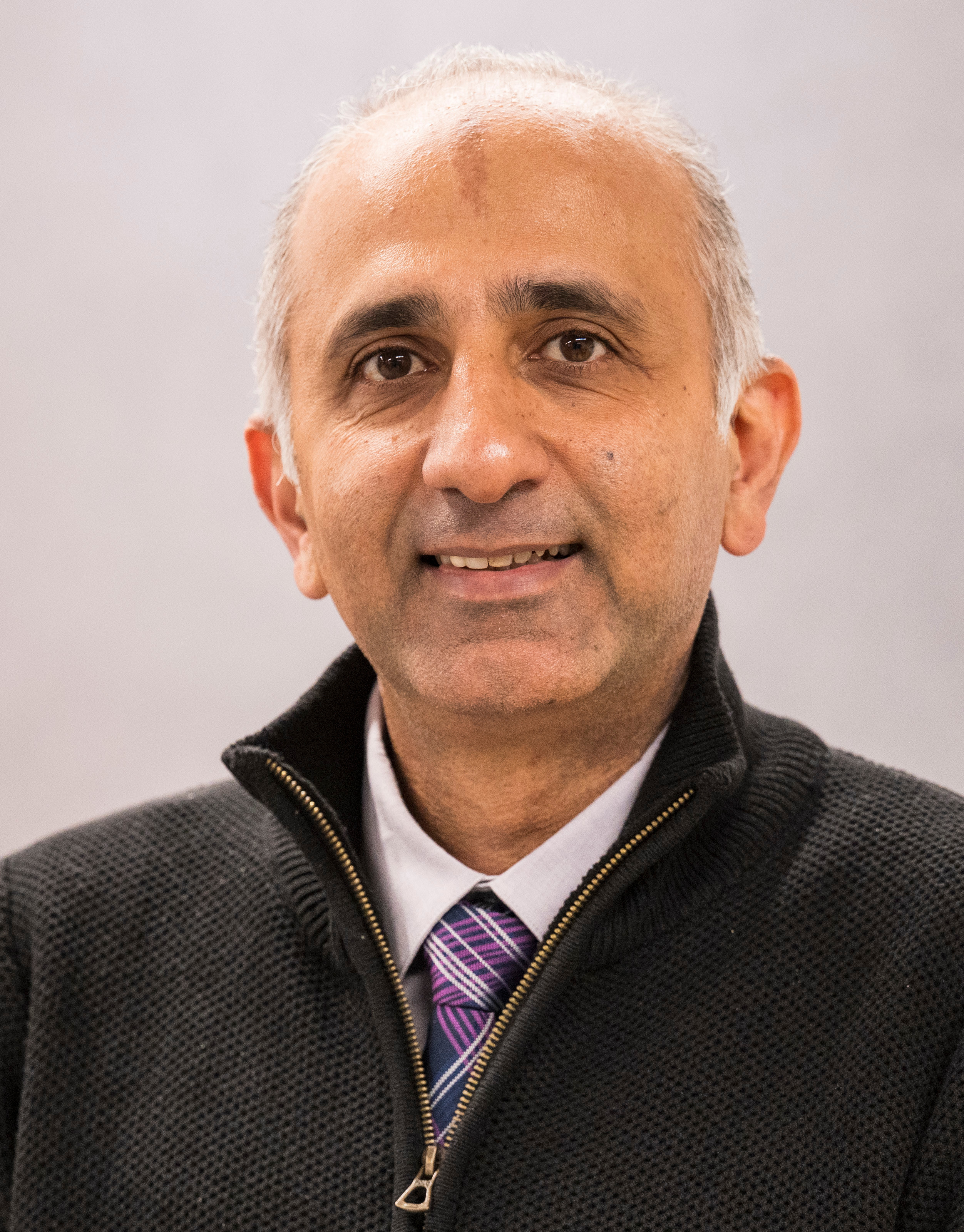}}]{Saber Fallah}
is Professor of Safe AI and Autonomy and the Director of CAV-Lab at the University of Surrey. He is an internationally recognized expert in the field of Safe AI and Autonomous Systems, focusing on the application of these technologies in self-driving cars and autonomous robotic systems. He leads pioneering research that seeks to ensure the safety, reliability, and trustworthiness of AI systems used in autonomous systems. His work adopts a strong interdisciplinary approach, bridging computer science, engineering, ethics, and public policy to address the challenges of integrating AI into autonomous technologies. He fosters collaboration between industry, government, academia, and local communities to ensure that self-driving vehicle development is technologically advanced, socially responsible, and aligned with public interests.
\end{IEEEbiography}

\end{document}